\documentclass[a4paper]{llncs}

\usepackage{times} % Please do not comment this
\usepackage{epsfig}
\usepackage{graphicx}
\usepackage{subfigure}
\usepackage{amsmath}
\usepackage{amssymb}
\usepackage{algorithm}
\usepackage{algorithmic}
\usepackage{defs}

\begin{document}

\pagestyle{empty}

\mainmatter

\title{Exploiting the Statistics of Learning and Inference}

\titlerunning{Exploiting the Statistics of Learning and Inference}

\author{Max Welling} 

\authorrunning{Max Welling}

\institute{Institute for Informatics\\
University of Amsterdam\\
Science Park 904, Amsterdam, Netherlands\\
\email{m.welling@uva.nl}}

\maketitle

\begin{abstract}
When dealing with datasets containing a billion instances or with simulations that require a supercomputer to execute, computational resources become part of the equation. We can improve the efficiency of learning and inference by exploiting their inherent statistical nature. We propose algorithms that exploit the redundancy of data relative to a model by subsampling data-cases for every update and reasoning about the uncertainty created in this process. In the context of learning we propose to test for the probability that a stochastically  estimated gradient points more than 180 degrees in the wrong direction. In the context of MCMC sampling we use stochastic gradients to improve the efficiency of MCMC updates, and hypothesis tests based on adaptive mini-batches to decide whether to accept or reject a proposed parameter update. Finally, we argue that in the context of likelihood free MCMC one needs to store all the information revealed by all simulations, for instance in a Gaussian process. We conclude that Bayesian methods will remain to play a crucial role in the era of big data and big simulations, but only if we overcome a number of computational challenges. 
\end{abstract}

\section{Statistical Learning}

When we learn a parametric model from data we extract the useful information from the data and store it in the parameter-values of the model. Naive algorithms learn all decimal places of the model parameters (up to machine precision) by optimizing some cost function (e.g. the log-likelihood). If the model has a large capacity to store information this might lead to overfitting. Regularization typically avoids parameters to become too large, preventing the learning algorithm to store information in the most significant bits of the parameters. Bayesian methods determine a posterior distribution over parameters, where the prior usually prevents the parameters from becoming too large (similar to regularization) and the integration over the parameters weighted by the posterior effectively destroys the information in the insignificant decimal places of the parameter values. More data usually implies that more bits of our parameters are recruited to store information.

In the context of big data, there is the general perception that A) learning is computationally more expensive implying that given a fixed amount of computational resources it takes longer to train good predictive models and B) that with more data overfitting is becoming less of a concern. However, both statements need not necessarily be true. To see why A) may not be true, one can simply imagine subsampling the large dataset into a smaller dataset and train a model model fast on the smaller dataset. Any algorithm that takes much longer to reach the same predictive power as this naive subsampling approach clearly does something wrong if one cares at all about \emph{the amount of learning per unit time}. Before the advent of big data one may not have cared much about reaching the optimal possible predictive performance given some fixed amount of time (or more generally any amount of time unknown in advance).  However, one can no longer afford this luxury when there is so much data that most algorithms will simply not completely converge within any reasonable amount of time. In other words, we should be interested in algorithms that when interrupted at any arbitrary time should be optimally predictive, and not with algorithms that perform well only after a very long time. A typical example of the former class of algorithms is "stochastic gradient descent" (SGD), while typical examples of the latter are "batch learning" and standard Markov chain Monte Carlo sampling algorithms (MCMC). SGD can be expected to do as well as possible for any fixed amount of training time (when the annealing schedule for the stepsize is set wisely), while batch learning and MCMC might return disastrous results when the amount of compute time is limited. As an extreme example, for a very large data-set a batch gradient update might not have done a single update while SGD might have already arrived at a reasonable model. 

To see why B) is not necessarily true we only have to remember that while data volume grows exponentially, Moore's law also allows us to train models that grow exponentially in their capacity. In the field of deep learning, this is exactly what seems to be happening: Google and Yahoo! currently train models with billions of parameters. At the same time, the amount of \emph{predictive information} in data grows slower than the amount of Shannon information \cite{BialekNemenmanTishby01} namely as $N^\al$ with $\al<1$. This law of diminishing returns of data thus implies that our models are increasing their capacity faster than the amount of predictive information that we need to fill them with. The surprising conclusion is thus that regularization to prevent overfitting is increasingly important rather then increasingly irrelevant. We have seen some evidence of this recently when \cite{HintonandSrivastavaKrizhevskySutskeverSalakhutdinov12} introduced dropout in an attempt to combat overfitting for deep neural networks. It seems that currently our best models are the ones that overfit to some degree and require regularization or bagging to avoid the overfitting. If, in the context of big data, you are training models that do not straddle the boundary between under- and overfitting, then it is likely that by increasing the capacity of your model (assuming that you have the computational resources to do so) you can reach better predictive performance. 

We thus argue forcefully that computationally efficient Bayesian methods are becoming increasingly relevant in the big data era: they are relevant because our best high capacity models need them as a protection against overfitting and they need to be computationally efficient in order to deal with the large number of data cases involved. There are essentially two classes of big data Bayesian methods and both are based on stochastic minibatch updates: stochastic gradient variational Bayes \cite{HoffmanBleiBach10} and stochastic gradient MCMC \cite{WellingTeh11}. I will say more about the latter later in this paper. 

\section{Statistical Optimization}

There is an increasing tendency to cast learning as an optimization problem of some loss function. For example, an SVM is often taught as a quadratic program over Lagrange multipliers and neural network training is taught as an exercise in minimizing weights of some loss function defined on the output units of the network. New powerful tools from the``convex optimization" literature have encouraged this myopic view of learning to the point that some researchers now hold the view that every learning problem should be cast as a (preferably) \emph{convex} optimization problem. The tendency to view all learning problems as ``mere optimization'' which can be successfully attacked with the modern blessings of convex optimization ignores some of the unique properties of learning problems. In other words: learning can indeed be formulated as an optimization problem, but a rather special one that has important features which should not be ignored. 

So what are these special properties of learning problems that distinguish them from plain vanilla optimization? The main difference lies in the loss function being a function of the data and the data being a random draw from some underlying distribution $P$. It is thus useful to think of the loss function as a random entity itself, i.e. one that fluctuates under resampling of the data-items from $P$. One aspect of this randomness is well appreciated by most machine learning researchers, namely that it contains the information necessary to avoid overfitting. Most researchers understand that simply fitting a model by minimizing some loss based on the training data alone might lead to overfitting. The fluctuations caused by resampling the training data will cause the optimal parameter-values to fluctuate and hence determine a distribution of parameters rather than a single value (the frequentist equivalent of the posterior distribution).

It is less appreciated that taking the statistical properties of optimization into account during the entire learning process (and not just close to the point of convergence) can be very helpful to increase its computational efficiency. Perhaps the point can be made most forcefully by considering the extreme case of an \emph{infinite} dataset. Any learning procedure that uses all data at every iteration is doomed to not do anything at all! During the initial stages of learning, when we are trying to the determine the most significant digits of our parameters, the information in the data is highly redundant. In other words: most data items agree on how they recommend changing the parameter values and as a result, one only has to query a small subset of them to get reliable information on how to update parameters. In contrast, close to convergence, most data items disagree on how to change the parameters and the update direction will strongly depend on ``who you ask''. Stochastic gradient descent (SGD) exploits precisely this idea, but does not tie it directly to the statistical properties of the optimization problem, implying that the annealing schedule of the stepsize has to be set by hand. In \cite{BoylesKorattikaraRamananWelling11,KorrattikaraBoylesWellingKimPark11} we have proposed methods to increase the minibatch size (instead of decreasing the stepsize), based on statistical hypothesis tests that estimate the probability that a proposed update is more than 180 degrees in the wrong direction, leading to a learning procedure that auto-tunes its optimization hyper-parameters. These methods exploit the redundancy in data in relation to a partially trained model.

\section{Data Redundancy}

As discussed in the previous section, our proposed methods \cite{BoylesKorattikaraRamananWelling11,KorrattikaraBoylesWellingKimPark11} for speeding up learning algorithms are based on the notion that far away from convergence only few data-cases are needed to determine a reasonably accurate update. For instance, if we want to learn the mean of a one dimensional Normal distribution, and the current mean is still far away from the sample mean, then we only need to query a few data-cases to know the direction of the update. We may say that the \emph{information in the data relevant to the parameter update is highly redundant}. It is interesting to emphasize that \emph{redundancy is not a property of the data, but of the relation between the model and the data}. For instance, close to convergence, about half of the data-cases recommend to update the mean of a Normal distribution to go left, while the other half recommend to update it to go right. At the ML estimate, the redundancy is exactly zero.

We will try to catch this intuition in a new measure, namely the ``learning signal-to-noise ratio" (LSNR) that measures how much learning signal there is relative to the noise of resampling the data-items. We will see that in the initial stages of learning this LSNR is large which means that we can use a smaller random subset of the data to estimate the gradient direction of the loss function reliably. As learning proceeds, the LSNR gradually decreases below $1$ prompting us to increase the size of our minibatch. Eventually, the LSNR will drop below $1$ even when we are using all the data available to us. At this point in the learning process the learning signal is too faint to be useful, i.e. the gradient direction could easily swing 180 degrees if we would have used a different dataset of the same size. More parameter updates on the current dataset would thus lead to overfitting and aversely affect generalization performance. We can avoid this by simply terminating the parameter updates at that point or by switching to more sophisticated methods such as bagging \cite{Breiman96}. The proposed metric is inherently  frequentist because it reasons about resampling data-sets, however we believe similar metrics can de devised to monitor data redundancy for model parameters  in a Bayesian setting. 

Denote with $\bar{\bg} = \nN\grad_{\bta} \ell(\bx_i;\bta)$ where $\ell(\cdot)$ is a general objective function, $\cD=\{\bx_i\}$ denotes the data and $\bta$ denotes the parameters of the problem.  Furthermore, denote with $S$ the sample variance-covariance matrix of the gradients,
\be
S = \frac{1}{n-1}\sum_{i=1}^n (\grad_{\bta} \ell(\bx_i;\bta) - \bar{\bg})(\grad_{\bta} \ell(\bx_i;\bta) - \bar{\bg})^T
\ee
If $n$ is large enough, according to the central limit theorem, the covariance matrix of the average gradients is then given as $S_{\bar{\bg}}\approx \frac{1}{n}S$. 

We are interested in the value of the signal in the gradient, given by $\bar{\bg}$ relative to the noise we expect under resampling of the data-items, given by $S_{\bar{\bg}}$. We propose the following measure:
\be
LSNR_p = \frac{1}{p}\bar{\bg}^TS_{\bar{\bg}}^{-1}\bar{\bg} \approx \frac{n}{p}\bar{\bg}^TS^{-1}\bar{\bg}
\ee
where $p$ indicates the number of parameters. 

A useful property of this metric is that it is invariant under linear transformations of the form $\bg_i\ra M\bg_i$. In particular, it is therefore independent of the stepsize $\eta$ used in a gradient descent algorithm. We can determine the sampling distribution\footnote{To see why this is true, we first note that the sample mean and covariance converge in probability to their respective population values for large enough $n$.
\be
\bar{\bg} = \nN  \bx_i \stackrel{P}{\rightarrow}\bmu,~~~~~~~~~~~~S = \frac{1}{n-1}\sum_{i=1}^n  (\bx_i - \bar{\bx})(\bx_i - \bar{\bx})^T\stackrel{P}{\rightarrow}\Sig
\ee
This means that according to Slutsky's theorem we can replace $S\ra \Sig$ in the expression for $LSNR_p$ when $n\gg 1$. We furthermore notice that according the central limit theorem $\bar{\bg}\sim \cN[\bmu,\frac{1}{n}\Sigma]$. We can then finally transform to $\bnu = n^\ha\Sig^{-\ha}\bar{\bg}$ with $\bnu\sim\cN[n^\ha\Sig^{-\ha}\bmu,I]$ and note that $\bnu^T\bnu = \sum_{j=1}^p \nu_j^2 \sim \chi_p^2(n\bmu^T\Sig^{-1}\bmu)$, which is what we wanted to show.} of $LSNR_p$ when $n\gg 1$.

\paragraph{\bf Lemma:} The random variable $p\times LSNR_p$ is asymptotically ($n\gg 1$) distributed as,
\be
p\times LSNR_p = n\bar{\bg}^TS^{-1}\bar{\bg} \sim \chi^2_p(n\bmu^T\Sig^{-1}\bmu)
\ee
where $ \chi^2_p(n\bmu^T\Sig^{-1}\bmu)$ is a non-central $\chi^2$-distribution with $p$ degrees of freedom and location parameter $n\bmu^T\Sig^{-1}\bmu$. To obtain the distribution for $LSNR_p$ (as opposed to $p\times LSNR_p$) we use that $X/p \sim p P_X(pX)$.

The behavior of $P(LSNR_p)$ is exactly as expected. As we increase $n$ we expect that the mean of this distribution becomes more positive because the variance in the computation of the gradients is expected to decrease as $1/n$. Indeed, we see that the mean of the distribution, given by $\eE[LSNR_p]=\frac{n}{p}\bmu^T\Sig^{-1}\bmu + 1$, grows linearly in $n$. We can also compute the variance, which is given by the expression $\eV[LSNR_p]=\frac{2}{p^2}(p+2n\bmu^T\Sig^{-1}\bmu)$. We note that the variance is of the order of the mean implying we should always expect significant fluctuations in the $LSNR_p$ under resampling the dataset. 

Two opposing effects determine the value of the $LSNR_p$ during the execution of a learning algorithm. The first effect is the one mentioned in the previous paragraph, namely that with increasing mini-batchsize $n_t$ the $LSNR_p$ increases because the fluctuations of the gradient are tempered due to central limit tendencies. The second effect is determined by how close to convergence we are in the learning process. If we are far from convergence the average gradient is expected to be large, even for small mini-batches. However, very close to convergence the average gradient is almost zero (actually zero at convergence by definition), meaning that the LSNR is also very small. In fact, for every finite dataset there will always be a point before convergence where the signal is not big enough to overcome the sampling noise.

How should we then use this $LSNR_p$ metric inside a learning algorithm? We envision a procedure where we compute the $LSNR_p$ at every iteration of learning to monitor the signal-to-noise ratio of the learning gradient. This could be useful in batch mode in order to decide when to stop the parameter updates, or this can be computed for a mini-batch $\cB_t$ to decide when to increase the mini-batch size. To compute the $LSNR_p$ one first estimates the mean and covariance from their sample estimators. In other words, we first compute the empirical gradients for every data-case in our current minibatch, $\bg_i = \grad_\ta\ell(\bx_i)$. From this we compute the sample mean and sample covariance,
\begin{align}
&\bmu\approx \hat{\bar{\bg}} = \frac{1}{n_t}\sum_{i\in \cB_t}\grad_{\bta} \ell(\bx_i;\bta)\\
&\Sig\approx \hat{S} =\frac{1}{n_t-1}\sum_{i\in \cB_t}(\grad_{\bta} \ell(\bx_i;\bta) - \hat{\bar{\bg}})(\grad_{\bta} \ell(\bx_i;\bta) - \hat{\bar{\bg}})^T
\end{align}
where the hatted variables are computed on the actual dataset (e.g. they are not random variables but numbers), and $\cB_t$ is the minibatch used at iteration $t$. This mini-batch should \emph{not} be resampled at every iteration because the $LSNR_p$ decreases relative to a fixed mini-batch.

Next, we use these estimates to compute the non-centrality parameter of the $\chi^2$ distribution: $n_t\bmu^T\Sig^{-1}\bmu\approx n_t\hat{\bar{\bg}}^T\hat{S}^{-1} \hat{\bar{\bg}}$. We emphasize that this computation scales with $n_t$, the size of the current mini-batch, and is thus relatively cheap. Using these we can for instance compute a threshold for deciding when the signal to noise ratio is too small to make meaningful updates. We suggest the following criterion,
\be
P(LSNR_p < 1) > \de~~~~~\leftrightarrow~~~~~~CDF(LSNR_p = 1 ) > \de
\label{eqn:crit}
\ee
where $\de=0.5$ seems a reasonable threshold. 

If this criterion is met one of two things might happen: 1) either we need to increase the size of the minibatch used to determine the gradient along which we are making weight updates, or 2) in case there is no more data available we terminate updating parameters in order to avoid overfitting. 

\begin{figure}[t]
\centerline{\psfig{figure=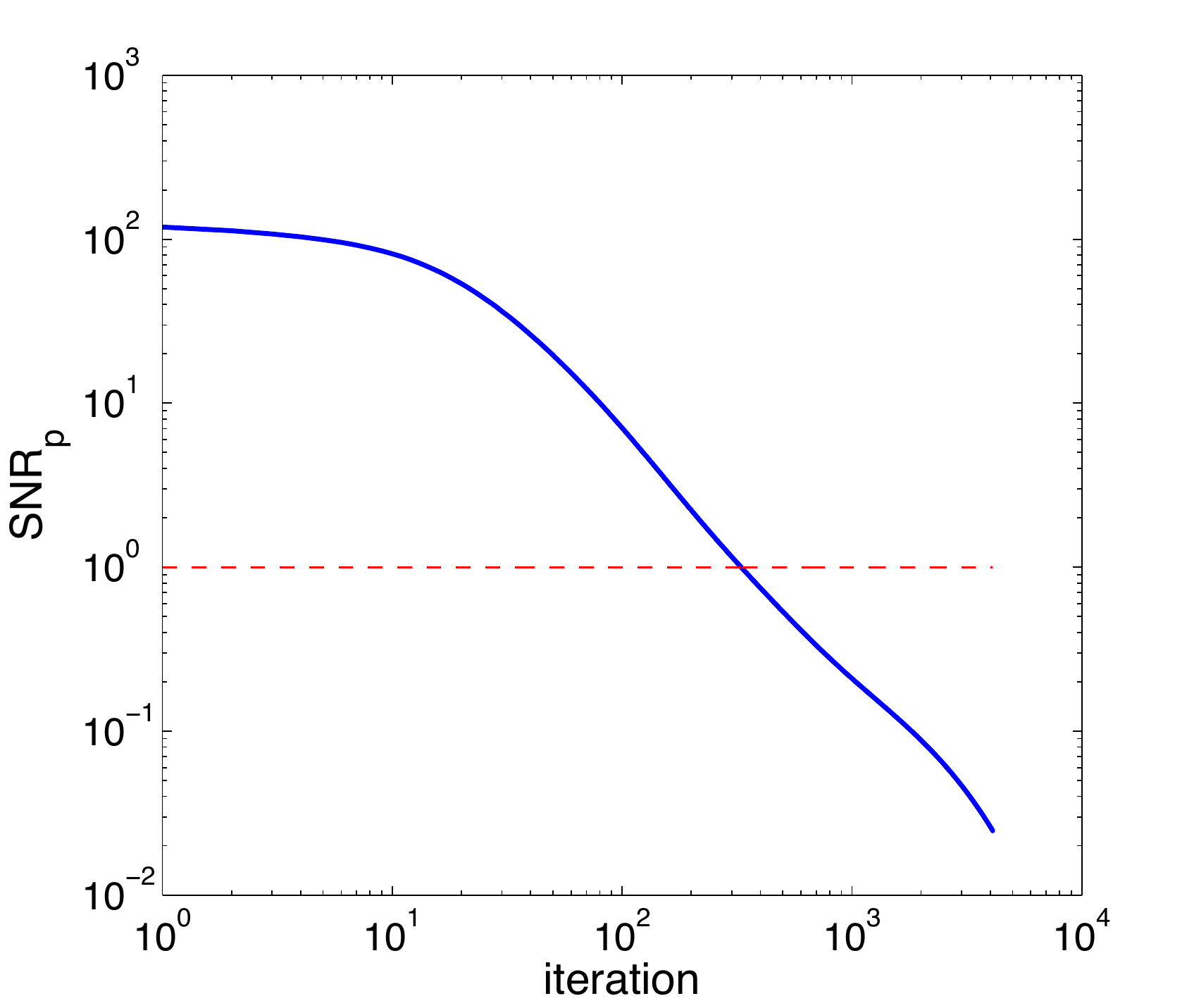,height=5cm}
\psfig{figure=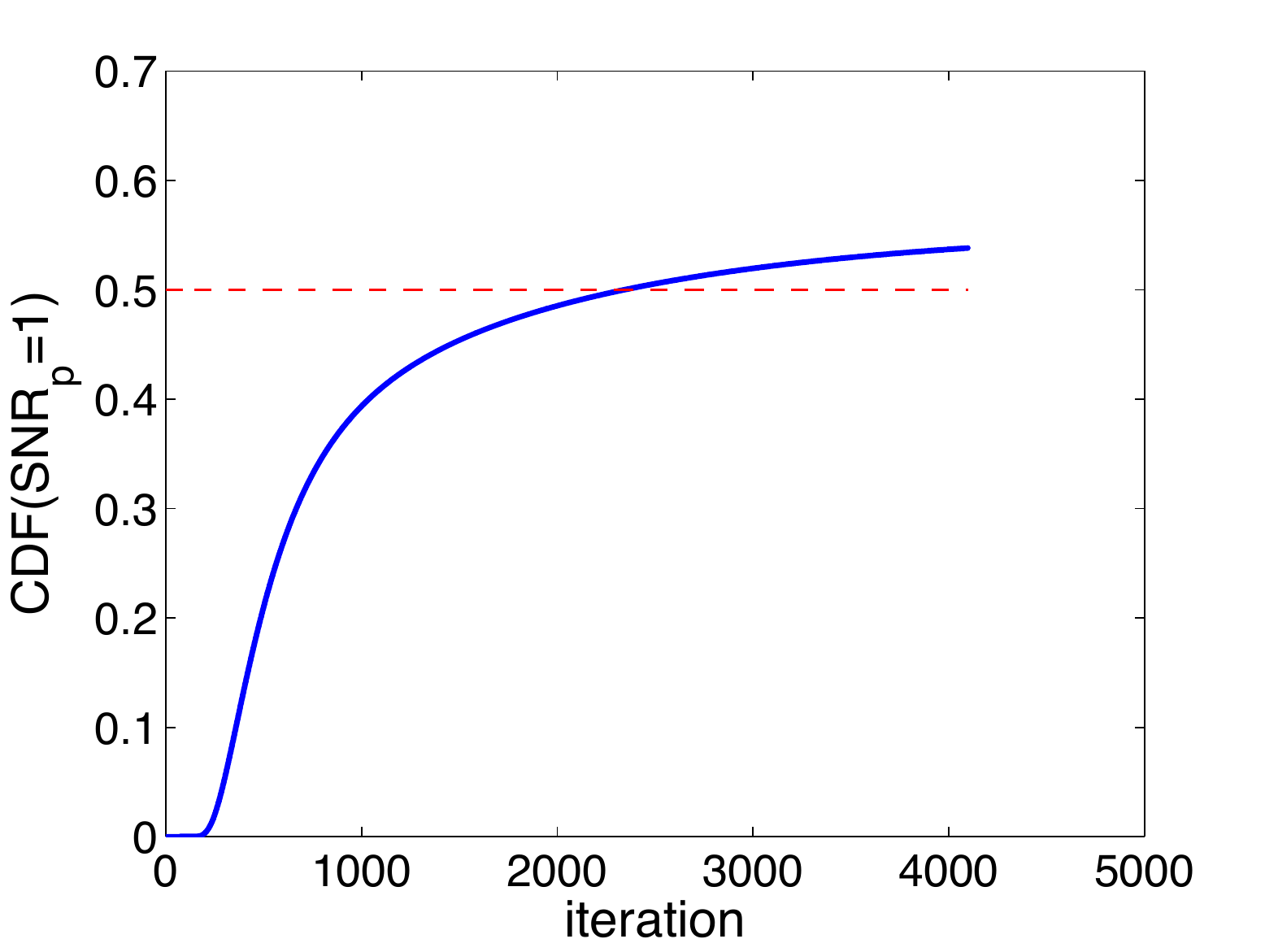,height=5cm}}
\caption{Left: Log-log plot of the value of the estimated $LSNR_p$ over time. Red dashed line indicates $LSNR_p=1$. Right: Value of $CDF(LSNR_p=1)$. Red dashed line indicates
threshold value of $0.5$.}
\label{fig:PDFCDFovertime}
\end{figure}
\begin{figure}[t]
\centerline{
\psfig{figure=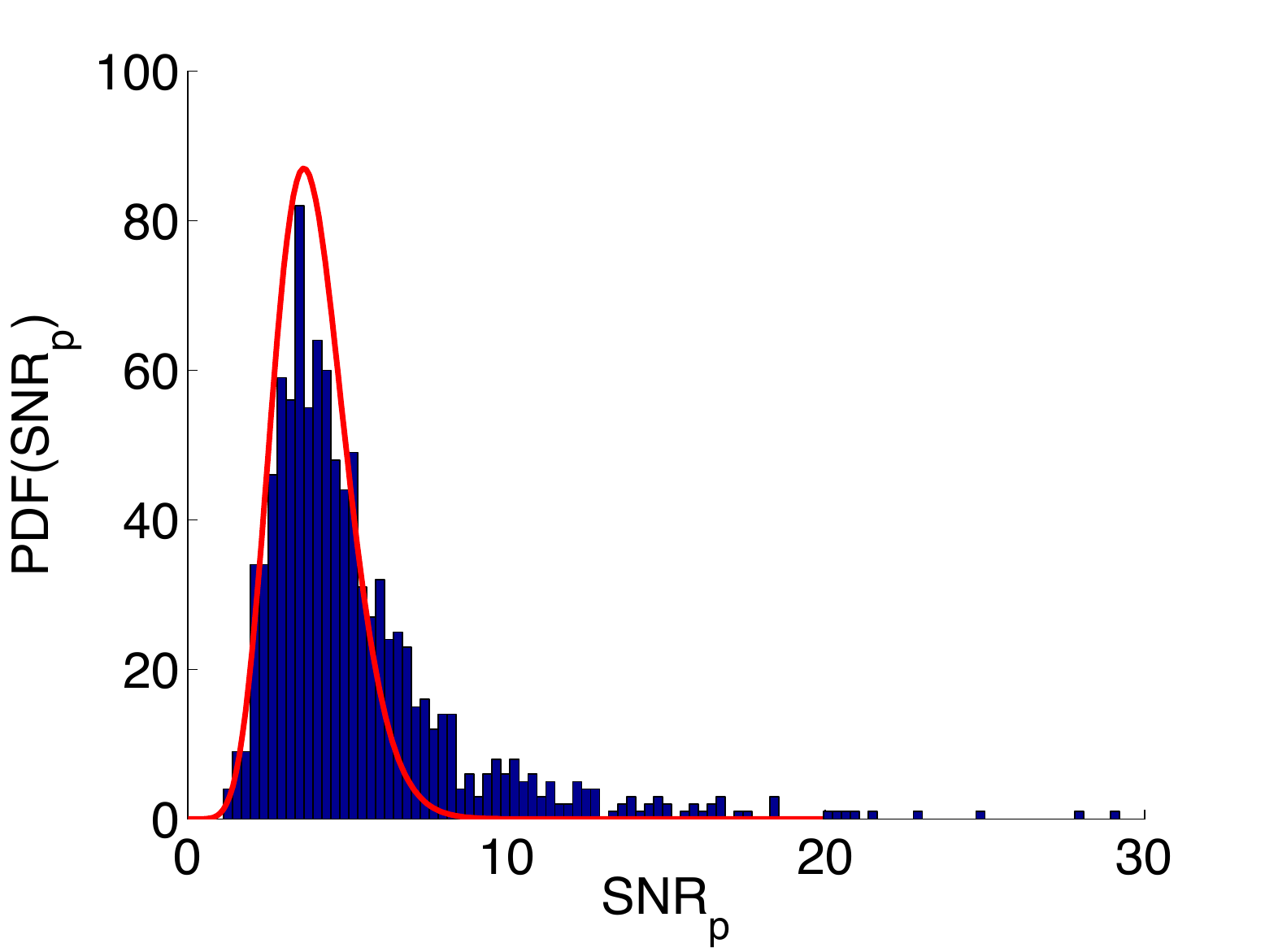,height=5cm}
\psfig{figure=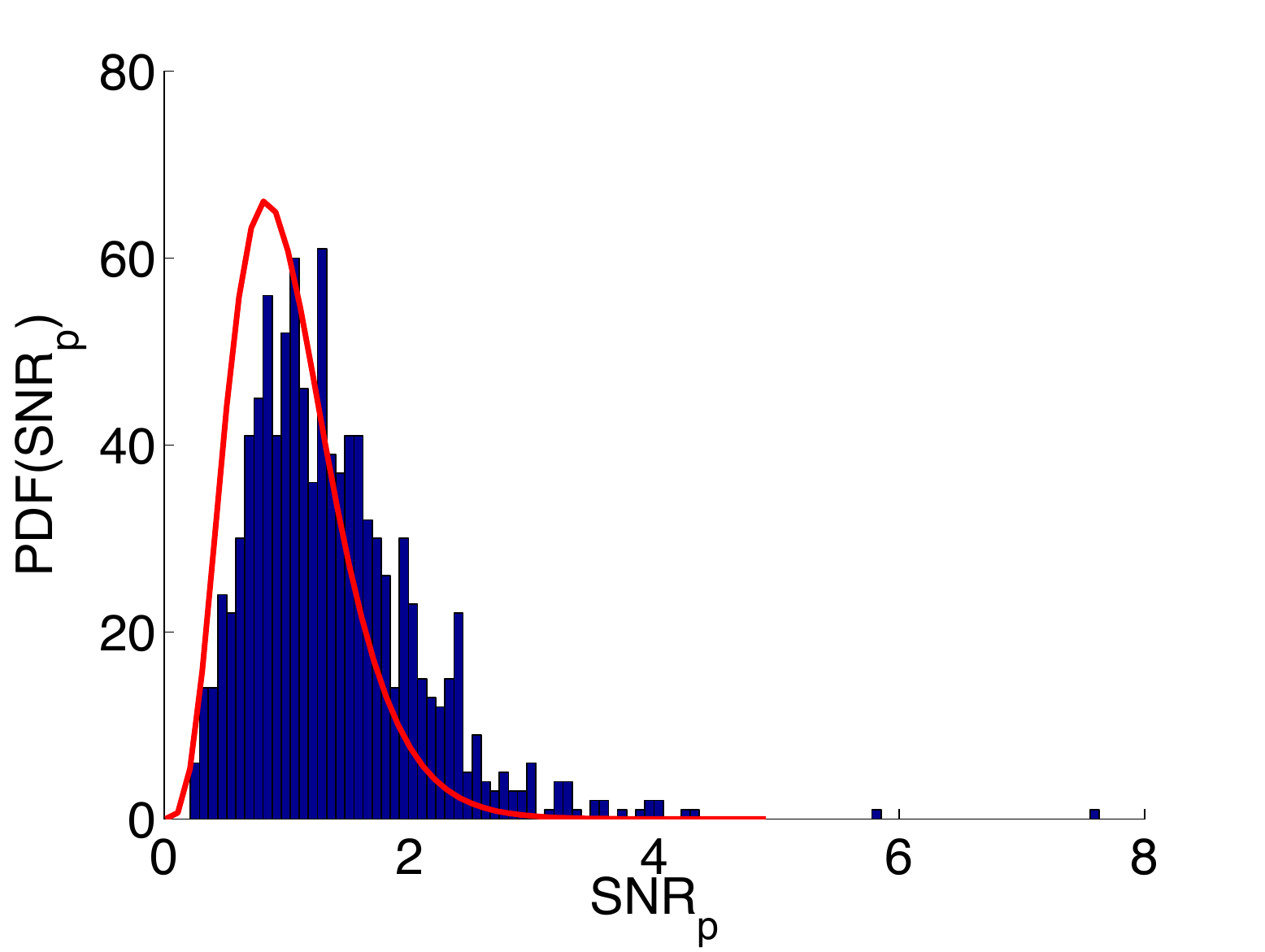,height=5cm}
}
\caption{Left: Non-central Chi-squared density for $LSNR_p$ at iteration $100$ of learning a logistic regression classifier. The overlaid histogram are the $LSNR_p$ values obtained from bootstrap samples of the original dataset. Density values are vertically scaled to compare more easily with the histogram. Right: same density and histogram but now at convergence.}
\label{fig:PDFSNR}
\end{figure}

To illustrate these ideas, we fit a logistic regression classifier (a.k.a. perceptron with logistic loss)  on 10 features of the spam dataset. We learn with all 3681 data-items at every iteration. In Figure \ref{fig:PDFCDFovertime} (left) we show the value of $LSNR_p$ as it evolves over time. We clearly see that it starts out large and decays rapidly towards zero. The red dashed line is the value of $LSNR_p=1$. Figure 1 (right) shows the value the cumulative distribution of $LSNR_p$ evaluated at $LSNR_p=1$. Thus, it represent the cumulative probability mass for the event $LSNR_p < 1$. Our proposed criterion \ref{eqn:crit} says that we should stop learning when it's value is larger than $0.5$, which is indicated with the red dashed line in the plot. Due to the asymmetry in the $\chi^2$-distribution this threshold is actually more conservative than $LSNR_p=1$ which is reached earlier. 

The plots of Figure \ref{fig:PDFSNR} show the $\chi^2$ distribution in the initial phases of learning (left) and close to convergence (right). We also show a histogram of $LSNR_p$ values obtained by first subsampling $1000$ bootstrap samples of size $n=3681$ from the original dataset and then computing the $LSNR_p$ value of those bootstrap samples. While bootstrap samples are not IID (independently identically distributed) they are generally accepted as a reasonable measure to assess variation under resampling of the dataset. We see that the fit of the $\chi^2$ is not perfect, but does provide a very reasonable approximation of the distribution of the $LSNR_p$ values obtained through bootstrapping. Note that the $LSNR_p$ is much larger on average when the model fit is still bad.

\section{Statistical MCMC Inference}
We now ask the question whether Bayesian posterior inference is ready to face the challenges of big data. Unfortunately, the answer is an unqualified no. Bayesian inference requires (an approximation of) the entire posterior distribution of parameters given data. Almost none of the standard posterior inference methods make use of statistical properties to improve the computational efficiency of inference (with the exception of \cite{HoffmanBleiBach10}). Let's consider the workhorse of Bayesian inference, MCMC sampling. Every MCMC algorithm that samples parameter instances from the posterior distribution $p(\bta|\cD)$ has to consider all data-items at every iteration. Thus, for the imaginary infinite dataset this algorithm comes to a grinding halt even before it starts. Even though for an infinite dataset the posterior consists of a single point (the ``maximum a posteriori'' value) the MCMC sampler never gets even close. The reason is that every MCMC procedure starts with a ``burn in" phase which often looks like a form of noisy optimization. However, standard MCMC procedures enforce detailed balance right from the start, which causes them to spend an unreasonably long time to finish burning in. For very large datasets, this might imply that MCMC procedures are likely not to even finish burning in, rendering them effectively useless in this context. 

But even close to convergence (after burn-in has finished), one can also benefit from the view proclaimed in the previous sections, namely that one should care about the best possible performance in a given (possibly unknown) amount of compute time. Traditional MCMC procedures assume that their chains can finish burning in and that a sufficient number of samples can be obtained to reduce the sampling variance thereafter. In the context of big data this might simply not be true anymore and it might become beneficial to design samplers that acquire samples faster but from an approximation of the true posterior (at least initially). By sampling faster, one can obtain a larger number of samples in a given amount of time and thus reduce sampling error due to variance. 

The total error can be expressed as the "risk" (the expected difference between the estimated average value of some function and its true average value). The risk in turn can be decomposed into the sampling error (or variance), which disappears for an infinite number of samples, and the bias, which represents a systematic error that does not go away asymptotically. The traditional view in MCMC has always been that one has enough time to reduce the sampling error to an arbitrary small value. While this view was reasonable at times when the datasets contained a few hundred items, in the era of big data we need to acknowledge that we only have finite computational resources (a.k.a. CPU cycles) to obtain the lowest risk in a fixed amount of time. Under these circumstances and depending on how much compute time one has available, it might be wiser to use a biased sampler in order to generate more samples and thus reduce the error due to variance faster. It should be mentioned that machine learning practitioners have already embraced this view some time ago in the context of posterior inference through biased methods such as variational Bayesian inference \cite{BealGhahramani03,Attias00,NealHinton99}and expectation propagation \cite{Minka01}.

I will briefly discuss two recent attempts by me and my collaborators to negotiate this tradeoff between bias and variance. The first approximate MCMC sampler is based on Langevin dynamics (LD). LD simply boils down to batch gradient descent with a stepsize $\eps$ plus the addition of normally distributed noise with mean equal to $0$ and variance equal to $\eps$. The coupling of the noise variance and the stepsize leads to Brownian motion type behavior of this dynamical system in the limit $\eps\rightarrow 0$ with an equilibrium distribution equal to the posterior distribution. Instead of using a very small stepsize it is often more efficient to treat one step of Langevin dynamics with finite stepsize as a proposal distribution within an ordinary MCMC algorithm and use a Metropolis Hastings (MH) accept-reject step to sample form the correct distribution asymptotically. A natural thought is thus to replace the full (batch) gradient computation by a stochastic minibatch estimate of it. One can show \cite{WellingTeh11} that when the stepsize $\eps_t$ is annealed to zero using the same conditions as the ones imposed for stochastic gradient descent in order to guarantee its convergence, namely $\sum_t \eps_t = \infty$, $\sum_t \eps_t^2 < \infty$, as $T\ra\infty$ this dynamical system samples from the correct distribution. If we also ignore (for the moment) the MH correction then the updates of this ``stochastic gradient Langevin dynamics" (SGLD) are $\cO(n)$ instead of $\cO(N)$ with $n<<N$ the size of the stochastically chosen minibatch. In practice, we do not anneal the stepsize all the way to zero because in this limit the sampler is not mixing anymore. By stopping at a finite value of the stepsize we thus accept a certain bias in the equilibrium distribution at the benefit of being able to sample much faster. 

SGLD has a number of interesting properties that make it an ideal candidate for large scale Bayesian posterior inference. The first observation is that for large stepsizes the noise induced by subsampling dominates the injected Gaussian noise because the subsampling noise has a variance $\cO(\eps^2)$ while the injected noise has a variance $\cO(\eps)$. For large stepsize, the algorithm thus effectively acts as stochastic gradient descent (SGD). However, when the stepsize gets smaller, the injected noise starts to dominate implying that the algorithm behaves as Langevin dynamics. If the annealing is thus done carefully, the algorithm switches from an efficient optimization algorithm into a posterior sampling algorithm. This effect can be further enhanced by using the empirical inverse Fisher information as a preconditioning matrix turning gradient descent into a form of stochastic Fisher scoring \cite{AhnKorattikaraWelling12}. 

A second property, namely that it only needs a small subset of data to generate samples, makes SGLD ideally suited as a distributed sampling algorithm. When a dataset is too large to store on a single server and the data is therefore distributed among multiple servers, traditional MCMC algorithms typically require these servers to communicate information for every sample generated. However, SGLD is able to make multiple parameter updates per server by subsampling the data on a single server without any communication \cite{AhnShahbabaWelling14}. This flexibility allows one to avoid servers having to sit idle and wait for other servers to finish their computation, even when servers store different amounts of data or have different processing speeds. The trick is to let all servers compute for the same amount of time before communicating their last parameter values to another randomly picked server, but to compensate potential bias due to unequal data volume or processing speed by adapting their relative stepsizes.

Omitting the MH accept step can result in a strong bias if there are regions where the probability drops very quickly. SGLD can step into these regions without being rejected and then due to a very large gradient get ``catapulted'' out. Motivated by this issue we investigated if we can design MCMC algorithms with MH steps that only depend on a small subset of the data \cite{KorattikaraChenWelling14} (see also \cite{BaDoHo14}). Naturally, without inspecting all the data at every iteration we will need to allow some errors to occur, but the question is if the number of errors can be controlled. To achieve this one can reformulate the MH test as a sequential hypothesis test. Given the uniform random variable $u$ that is used to make the final decision on whether to accept or reject, we are testing the sign of the mean of differences in log-likelihood between two parameter values:
\be
\mu = \frac{1}{N} \sum_{i=1}^N \left(\ell(\bx_i;\bta') - \ell(\bx_i;\bta_t)\right) > \mu_0 ?
\ee
with $\mu_0$ some constant that depends on $u$, the proposal distribution and the prior. Since this is a sum of terms we expect that this quantity is normally distributed when the number of terms in the sum is large enough. We start with an initial test with a small minibatch of size $n$ and compute the probability that $\mu<\mu_0$ or $\mu>\mu_0$. If we have enough confidence in either of these two possibilities we can accept or reject the new parameter value with a pre-specified confidence level. If neither is the case, we need to increase our minibatch size in order to increase the confidence level. By sampling new data-cases without replacement the standard deviation of distribution for $\mu$ behaves as,
\be
\sg(\mu)\propto \frac{1}{{\sqrt{n}}}\sqrt{1-\frac{n}{N}}
\ee 
where the first factor is the usual $\frac{1}{{\sqrt{n}}}$ behavior of the standard deviation of an average and the second term results from the fact that we are sampling without replacement. The second term is important in the sense that the standard deviation converges to $0$ when $n$ approaches $N$, implying that in that case an accept or reject decision will be made with 100\% confidence, consistent with a normal MH step.

An uncertain MH step is yet another way to negotiate the bias-variance tradeoff. Just like the stepsize in SGLD was a knob that traded off the error due to bias with error due to variance, here the confidence threshold for making an accept or reject decision acts as a bias-variance knob. If we make decisions quickly based on small minibatches of data we are bound to make more mistakes (accept moves that should be rejected and vice versa), but can collect more samples in a given amount of time (high bias, small variance). The algorithm reverts back to the standard MH-MCMC algorithm when all bias is eliminated and all error is due to variance.

We anticipate that many more algorithms can be developed along these lines. Indeed, \cite{DuboisKorattikaraWellingSmyth14} have developed a slice sampler variant based on these principles while \cite{ChenFoxGuestrin14} have developed a minibatch Hamiltonian Monte Carlo algorithm.

\section{Big Simulations}

Leaving the scene of big data we will now discuss the realm of big simulations. Just like the big data setting, Bayesian inference can play a crucial role in this context but is challenged computationally to its very limit. What then do we mean with "big simulations"?

Outside of machine learning, where we are happily playing with our graphical models and neural networks, scientists such as astronomers, meteorologists, seismologists, biologists and so on express their expert knowledge in terms of extremely complex simulators. To give an example of how extremely complex these simulations can get: the SCEC ``Shakeout simulation" of 360 minutes of earthquake simulation along the San Andreas fault in California took 220 teraflops per second for 24 hours on a NCCS Jaguar supercomputer with 223,074 cores \cite{CuiOlsenJordanLeeZhouSmallRotenElyChourasiaLevesqueDayMaechling10}. The typical simulator has a number of free parameters to be tuned in order to fit the observations well. Invariably the approach taken is to perform a grid search, systematically scanning through parameter values and comparing the outcome of the simulation with observations. Clearly, for very expensive simulations and high dimensional parameter spaces this is a hopeless endeavor. The recent advances in Bayesian optimization \cite{SnoekLarochelleAdams12} should prove to be a huge help in replacing grid search by a smarter form of parameter exploration However, scientists want and need to know more about their simulator than just a single optimal parameter setting. Questions such as: "Does my simulator fit the data well because I have captured the true underlying physical process or because my model is so flexible that I am overfitting." can not be answered by identifying a single optimal parameter setting. Instead, one needs to study the posterior predictive distributions, i.e.
\be
p(X|\cD) = \int \td \bta P(X|\bta)P(\bta|\cD)
\ee
and compare them with the actual observations. For the above expression it should be noted that $P(X|\bta)$ is not available as an analytic expression but only indirectly through a simulator. Precisely because of this reason it is also difficult to compute the posterior distribution $P(\bta|\cD)$, for which a special class of likelihood-free MCMC algorithms has been developed generally known as "Approximate Bayesian Computation" (ABC).

While the field of ABC is relatively well matured, we believe it is not ready to face the computational challenges of the very complex simulations that require supercomputers to execute. We like to emphasize the importance of this issue. While Moore's law allows scientists to design increasingly complex simulations, if they don't have the statistical tools to reliably verify if their models describe the truth then scientific progress comes to a grinding halt. The challenge for the computational statisticians and the machine learners is to turn this state of affairs around. 

A typical ABC algorithm works as follows. Like in ordinary MCMC we propose a new parameter value from some proposal $Q(\bta'|\bta_t)$. But due to the lack of an expression for the likelihood we cannot directly compute the probability of accepting this proposed parameter value. Instead, we will conduct a number of simulations and compare these simulations with the observations. Then, we accept the new parameter value if the simulations of the new model $\bta'$ are not much worse than the ones generated form the old model $\bta_t$ using the usual MH mechanism. There are various ways to compare the two collections of simulations but all require one to perform multiple simulations for every proposed new parameter value. For complex simulators, this is simply too expensive to be of use. 

How can this conundrum be resolved? In our opinion the answer lies in the fact that traditional ABC methods do not make efficient use of the information extracted from every simulation. Imagine our MCMC sampler would, after some detour, arrive back at a parameter value very close to the one where we just did an expensive simulation. Assuming some smoothness in the (unknown) likelihood we should be able to reuse the old simulation in order to make an informed decision about this new parameter value. Thus, we should store the information for all previous simulations and reuse them to make MH accept/reject decisions in the future. By learning a surrogate function of the unknown likelihood surface we can at some point avoid simulations altogether. The situation is somewhat similar to the approximate MH step introduced in the context of big data in that we will need to estimate the uncertainty in MH decisions and request more simulations only when the confidence in the MH decision is too low. 

In  \cite{MeedsWelling13} we proposed the following procedure based on the notion of a synthetic likelihood \cite{Wood10}. The ``naive'' synthetic likelihood procedure generates a number of samples at every iteration (for every parameter value) and fits a normal distribution through these samples. It then computes the probability (likelihood) of the observations under this Gaussian model inside the MH step. Our Gaussian Process Surrogate (GPS) procedure updates a Gaussian process (GP) instead of recalculating the Gaussian likelihood for every proposed parameter value. In that way, the information of all simulations is stored in the GP and for every pair of old and new parameters $(\bta_t,\bta')$ we can compute a full probability distribution over all the observed sufficient statistics $P(X|\bta)$, \emph{including the uncertainty in these distributions}. If this uncertainty is too high our accept or reject decisions are are too uncertain triggering a request for more simulations. At what parameter values these new simulations are conducted is entirely at the discretion of the algorithm and need not not coincide with $(\bta_t,\bta')$. Clearly, the longer we sample, the more we reduce the uncertainties in our GP and the less simulations will be requested in the future.

There are a number of additional dimensions along which this process can be further optimized. It would for instance be nice to use a procedure akin to Bayesian optimization to propose new parameter values to be examined. However, guaranteeing that this chain converges to the correct distributions is difficult because detailed balance is lost. Learning the GP's hyper-parameters and extending the GP to deal with potential heteroscedasticity are also important directions of further refinement.

\section{Conclusions}
Our claim is that in the context of large datasets and complex simulations, \emph{it is imperative that we leverage the statistical properties of the learning and inference process}. Intuitively, we want to maximize the amount of information that we transfer from data or simulations to parameters per unit of computation.  We are thus drawn to the conclusion that computation must be an essential ingredient of the overall objective. The traditional view of learning and inference could be maintained because the datasets of "the old days'' were small enough and the simulations of "the old days" were simple enough, so that relatively expensive optimization or inference procedures would still converge in a matter of hours. However, when one is faced with 1 billion data-cases or a simulation that runs 24 hours on a supercomputer the equation changes dramatically and one is forced to rethink traditional inference methods such as MCMC. We predict that Bayesian methods will remain to play an important role in an era where data volume and simulation complexity grow exponentially only if we manage to overcome a number of computational challenges.

\bibliographystyle{plain}
\bibliography{../Refs}

\begin{thebibliography}{10}

\bibitem{AhnKorattikaraWelling12}
S.~Ahn, A.~Korattikara, and M.~Welling.
\newblock Bayesian posterior sampling via stochastic gradient fisher scoring.
\newblock In {\em International Conference on Machine Learning}, pages
  1591--1598, 2012.

\bibitem{AhnShahbabaWelling14}
S.~Ahn, B.~Shahbaba, and M.~Welling.
\newblock Distributed stochastic gradient mcmc.
\newblock Technical report, University of California Irvine, 2014.

\bibitem{Attias00}
H.~Attias.
\newblock A {Bayesian} framework for graphical models.
\newblock In {\em Advances in Neural Information Processing Systems -- NIPS},
  volume~12, 2000.

\bibitem{BaDoHo14}
R.~Bardenet, A.~Doucet, and C.~Holmes.
\newblock Towards scaling up mcmc: an adaptive subsampling approach.
\newblock In {\em International Conference on Machine Learning (ICML)}, 2014.

\bibitem{BealGhahramani03}
M.J. Beal and Z.~Ghahramani.
\newblock The variational {Bayesian} {EM} algorithm for incomplete data: with
  application to scoring graphical model structures.
\newblock In {\em Bayesian Statistics}, pages 453--464. Oxford University
  Press, 2003.

\bibitem{BialekNemenmanTishby01}
W.~Bialek, I.~Nemenman, and N.~Tishby.
\newblock Predictability, complexity and learning.
\newblock {\em Neural Computation}, 13:2409--2463, 2001.

\bibitem{BoylesKorattikaraRamananWelling11}
L.~Boyles, A.~Korattikara, D.~Ramanan, and M.~Welling.
\newblock Statistical tests for optimization efficiency.
\newblock In {\em Neural Information Processing Systems}, 2011.

\bibitem{Breiman96}
Leo Breiman.
\newblock Bagging predictors.
\newblock {\em Machine Learning}, 24(2):123--140, 1996.

\bibitem{ChenFoxGuestrin14}
T.~Chen, E.B. Fox, and C.~Guestrin.
\newblock Stochastic gradient hamiltonian monte carlo.
\newblock Technical report, University of Washington, 2014.

\bibitem{CuiOlsenJordanLeeZhouSmallRotenElyChourasiaLevesqueDayMaechling10}
Y.~Cui, K.B. Olsen, T.H. Jordan, K.~Lee, J.~Zhou, P.~Small, D.~Roten, G.~Ely,
  D.K. Panda, A.~Chourasia, J.~Levesque, S.M. Day, and P.~Maechling.
\newblock Scalable earthquake simulation on petascale supercomputers.
\newblock In {\em International Conference for High Performance Computing,
  Networking, Storage and Analysis}.

\bibitem{DuboisKorattikaraWellingSmyth14}
C.~DuBois, A.~Korattikara, M.~Welling, and P.~Smyth.
\newblock Approximate slice sampling for bayesian posterior inference.
\newblock In {\em Artificial Intelligence and Statistics}, 2014.

\bibitem{HintonandSrivastavaKrizhevskySutskeverSalakhutdinov12}
G.E. Hinton, N.~Srivastava, A.~Krizhevsky, I.~Sutskever, and R.~Salakhutdinov.
\newblock Improving neural networks by preventing co-adaptation of feature
  detectors.
\newblock Technical Report abs/1207.0580, University of Toronto, 2012.

\bibitem{HoffmanBleiBach10}
M.~Hoffman, D.~Blei, and F.~Bach.
\newblock Online learning for latent dirichlet allocation.
\newblock In {\em Neural Information Processing Systems}, pages 856--864, 2010.

\bibitem{KorrattikaraBoylesWellingKimPark11}
A.~Korattikara, L.~Boyles, M.~Welling, J.~Kim, and H.~Park.
\newblock Statistical optimization of non-negative matrix factorization.
\newblock In {\em Artificial Intelligence and Statistics}, pages 128--136,
  2011.

\bibitem{KorattikaraChenWelling14}
A.~Korattikara, Y.~Chen, and M.~Welling.
\newblock Austerity in mcmc land: Cutting the metropolis-hastings budget.
\newblock In {\em International Conference on Machine Learning (ICML)}, 2014.

\bibitem{MeedsWelling13}
T.~Meeds and M.~Welling.
\newblock Gps-abc: Gaussian process surrogate approximate bayesian computation.
\newblock arXiv 1401.2838, University of Amsterdam, 2013.

\bibitem{Minka01}
T.~Minka.
\newblock Expectation propagation for approximate {B}ayesian inference.
\newblock In {\em Proc. of the Conf. on Uncertainty in Artificial
  Intelligence}, pages 362--369, 2001.

\bibitem{NealHinton99}
R.M. Neal and G.E. Hinton.
\newblock A view of the {EM} algorithm that justifies incremental, sparse and
  other variants.
\newblock {\em Learning in Graphical Models}, pages 355--368, 1999.

\bibitem{SnoekLarochelleAdams12}
J.~Snoek, H.~Larochelle, and R.~A. Adams.
\newblock Practi- cal bayesian optimization of machine learning algorithms.
\newblock In {\em Neural Information Processing Systems,}, 2012.

\bibitem{WellingTeh11}
M.~Welling and Y.W. Teh.
\newblock Bayesian learning via stochastic gradient langevin dynamics.
\newblock In {\em Proceedings of the 28th International Conference on Machine
  Learning (ICML)}, pages 681--688, 2011.

\bibitem{Wood10}
S.N. Wood.
\newblock Statistical inference for noisy nonlinear ecological dynamic systems.
\newblock {\em Nature}, 466(7310):1102--1104, 2010.

\end{thebibliography}
\end{document}